\documentclass[11pt]{article}

\usepackage[margin=1in]{geometry}
\usepackage{amsmath}
\usepackage{amssymb}
\usepackage{graphicx}
\usepackage{authblk}
\IfFileExists{booktabs.sty}{%
  \usepackage{booktabs}%
}{%
  \newcommand{\toprule}{\hline\hline}
  \newcommand{\midrule}{\hline}
  \newcommand{\bottomrule}{\hline\hline}
}
\IfFileExists{xcolor.sty}{\usepackage{xcolor}}{\usepackage{color}}
\usepackage[numbers,sort&compress]{natbib}
\newif\ifhyperrefavailable
\IfFileExists{etoolbox.sty}{%
  \hyperrefavailabletrue
  \usepackage{hyperref}%
}{%
  \hyperrefavailablefalse
  \usepackage{url}%
}
\usepackage{array}
\definecolor{linkblue}{RGB}{0,72,122}
\definecolor{todored}{RGB}{150,20,20}
\ifhyperrefavailable
  \hypersetup{
    colorlinks=true,
    linkcolor=linkblue,
    citecolor=linkblue,
    urlcolor=linkblue,
    pdfauthor={Authors to be inserted},
    pdftitle={Neural Operators for Immersed-Boundary Swimmers Locomotion}
  }
\else
  \ifdefined
  \fi
\fi

\newcommand{\vect}[1]{\boldsymbol{#1}}
\newcommand{\Rey}{\mathrm{Re}}

\title{\textbf{Neural Operators for Immersed-Boundary Soft Swimmers Locomotion}}
\author[1]{Mohammad Sadegh Eshaghi\thanks{Corresponding author. Email: eshaghi.khanghah@iop.uni-hannover.de}} 
\author[2]{Yizheng Wang} 
\author[1]{Navid Valizadeh \thanks{Corresponding author. Email: navid.valizadeh@gmail.com }} 
\author[1]{Xiaoying Zhuang} 
\author[3]{Timon Rabczuk \thanks{Corresponding author. Email: timon.rabczuk@uni-weimar.de}}

\affil[1]{Chair of Computational Science and Simulation Technology, Institute of Photonics, Department of Mathematics and Physics, Leibniz University Hannover, Germany}
\affil[2]{Department of Engineering Mechanics, Tsinghua University, Beijing, China}
\affil[3]{Institute of Structural Mechanics, Bauhaus-Universität Weimar, Germany}

\date{}

\begin{document}

\maketitle

\begin{abstract}
High-fidelity immersed-boundary simulation resolves the coupled motion of a
deforming swimmer and its surrounding flow, but the resulting cost limits
repeated evaluations for engineering design, parameter studies, and control.
We develop neural-operator surrogates for temporal prediction of the
hydrodynamic fields generated by planar and volumetric eel swimmers. The
surrogates are trained on regular-grid fields exported from adaptive
fluid--structure simulations and are conditioned on swimmer geometry and
Reynolds number. The planar model jointly predicts two velocity components,
scalar vorticity, and pressure. On five held-out high-Reynolds-number
trajectories, its full-domain global relative \(L^2\) error is \(3.51.\%\). The volumetric
formulation uses three target-specific models with a common multichannel input:
one model predicts three-dimensional velocity, one predicts vorticity, and one
predicts pressure. Their full-domain global relative \(L^2\) errors on five
held-out within-range trajectories are \(3.44\%\), \(5.58\%\), and \(19.2\%\).
Together, the results demonstrate the feasibility of field-resolved neural
surrogates for moving-boundary swimmer flows while identifying pressure
accuracy and physical consistency as priorities for further development.
\end{abstract}


\section{Introduction}
\label{sec:introduction}

High-fidelity simulation is a central tool for analyzing bio-inspired and soft
underwater swimmers. The hydrodynamic response depends on a moving and
deforming body, nonlinear wake dynamics, pressure loading, and viscous shear.
Resolving these interactions supports design-space exploration, inverse
identification, uncertainty analysis, and control, but repeated
fluid--structure interaction (FSI) calculations remain expensive.

Immersed-boundary methods avoid repeated body-fitted remeshing by coupling a
Lagrangian representation of the swimmer to an Eulerian fluid grid ~\cite{peskin2002ib}. Adaptive implementations such as IBAMR make detailed
moving-boundary simulations practical~\cite{griffith2007adaptive,
griffith2017hybrid,ibamrsoftware}, yet a parameter sweep or optimization loop
still requires many trajectories. A surrogate that returns distributed
velocity, vorticity, and pressure fields can retain substantially more
engineering information than a scalar performance predictor while reducing
the cost of repeated model queries.

Canonical mathematical and partial-differential-equation benchmarks have been
essential for developing and comparing neural operators
~\cite{li2021fno,lu2021deeponet,li2021pino,kovachki2023neuraloperator}.
Controlled benchmarks isolate approximation and optimization effects. A
complementary question is whether the same methods remain effective under the
moving geometry, coupled multiphysics, multichannel outputs, parameter
variation, and volumetric data costs of a high-fidelity engineering workflow.
Immersed-boundary swimmer locomotion provides such an application-scale
setting without diminishing the continuing value of canonical benchmarks.

Bio-inspired robots further motivate efficient predictive models. Soft robotic
fish have demonstrated untethered operation in natural aquatic environments
~\cite{katzschmann2018sofi}, and differentiable simulators have enabled
calibration and co-design of swimmer geometry, materials, and control
~\cite{ma2021diffaqua,zhang2022material,nava2022fast,lee2023aquarium}.
The present study asks whether neural operators
can reproduce the Eulerian fields of a high-fidelity immersed-boundary model.

We study planar and volumetric eel swimmers over Reynolds-number sweeps. The
planar network predicts velocity, scalar vorticity, and pressure jointly. The
volumetric case uses three separate target-specific networks for velocity,
vorticity, and pressure. A reference-state temporal mapping provides the
common accuracy assessment. 

The principal contributions are:
\begin{enumerate}
  \item a neural-operator surrogate formulation for
        immersed-boundary swimmer flows in both two and three dimensions,
        including moving-geometry and flow-parameter conditioning;
  \item quantitative, field-resolved assessment of velocity, vorticity, and
        pressure over held-out Reynolds-number trajectories, with explicit
        full-domain, fluid-region, component-wise, and physical-consistency
        metrics; and
  \item demonstration of neural field prediction for both planar and genuinely
        volumetric high-fidelity swimmer simulations, including
        parameter-dependent behavior and slice-resolved three-dimensional
        field reconstruction.
\end{enumerate}

\section{Related Work}
\label{sec:related}

\subsection{Bio-inspired swimmers and immersed-boundary simulation}

Soft swimmers combine compliant mechanics with distributed actuation, making
their hydrodynamics difficult to approximate with rigid-link or quasi-steady
models; Yasa et al.\ review the broader soft-robotics landscape
~\cite{yasa2023overview}. DiffAqua introduced a differentiable design pipeline
for underwater swimmers~\cite{ma2021diffaqua}; subsequent work combined
differentiable mechanics with learned or calibrated hydrodynamic models
~\cite{zhang2022material,nava2022fast}. Aquarium formulated a differentiable
two-dimensional FSI solver for robotics applications
~\cite{lee2023aquarium}. These studies show the value of efficient simulation
at several levels of fidelity.

The immersed-boundary method represents the fluid on an Eulerian grid and the
structure with Lagrangian coordinates, coupled through regularized delta
kernels~\cite{peskin2002ib}. Adaptive refinement concentrates resolution near
moving structures and wakes~\cite{griffith2007adaptive}. The unified
constraint-based formulation implemented in IBAMR accommodates rigid,
deforming, and elastic bodies within one mathematical framework
~\cite{bhalla2013unified}. This formulation supplies the simulation fields used
here.

\subsection{Neural operators and learned flow surrogates}

DeepONet represents nonlinear operators through branch and trunk networks
~\cite{lu2021deeponet}. The Fourier neural operator (FNO) instead learns
low-frequency spectral multipliers and has demonstrated parametric and
time-dependent field prediction on regular grids~\cite{li2021fno}.
The broader neural-operator framework formalizes maps between function spaces
and unifies several integral-kernel parameterizations
~\cite{kovachki2023neuraloperator}. Physics-informed neural operators add
equation-based constraints to paired data~\cite{li2021pino}, while
graph-based networks provide an alternative for irregular discretizations
~\cite{pfaff2021meshgraphnets,sanchezgonzalez2020learning}.

Geometry-aware variants extend operator learning beyond rectangular domains.
Geo-FNO learns coordinate deformations for general geometries
~\cite{li2023geofno}, and the geometry-informed neural operator targets
large-scale three-dimensional PDEs~\cite{li2023gino}. U-FNO combines Fourier
layers with a U-shaped architecture for three-dimensional multiphase flow
~\cite{wen2022ufno}, and learned components have also accelerated conventional CFD
~\cite{kochkov2021mlcfd}. These results motivate
moving-boundary, three-dimensional engineering applications, but swimmer
locomotion additionally couples prescribed body motion, wake transport,
pressure, geometry masks, and a broad parameter sweep.

Physics-aware and temporal extensions offer natural next steps. Variational
physics-informed neural operators enforce weak or energy forms
~\cite{eshaghi2024vino}; multi-head neural operators model multiple future
states with explicit temporal structure~\cite{eshaghi2026mhno}; and
neural-operator warm starts combine learned estimates with iterative solvers
~\cite{eshaghi2025nows}. The present work establishes a field-resolved FNO
baseline against which such developments can be evaluated.

\section{Problem Formulation}
\label{sec:problem}

\subsection{Immersed flow--structure dynamics}

Let \(\Omega_f\subset\mathbb{R}^{d}\), \(d\in\{2,3\}\), denote the Eulerian
fluid domain, and let \(\vect{X}(\vect{s},t)\) denote the Lagrangian swimmer
configuration. A schematic immersed-boundary system is
\begin{align}
 \rho\left(\frac{\partial \vect{u}}{\partial t}
 + \vect{u}\cdot\nabla\vect{u}\right)
 &= -\nabla p+\mu\nabla^2\vect{u}+\vect{f}_{\mathrm{IB}},
 \label{eq:momentum}\\
 \nabla\cdot\vect{u} &= 0,
 \label{eq:incompressibility}\\
 \vect{f}_{\mathrm{IB}}(\vect{x},t)
 &= \int_{\mathcal{B}}\vect{F}(\vect{s},t)
    \delta\!\left(\vect{x}-\vect{X}(\vect{s},t)\right)\,d\vect{s},
 \label{eq:spread}\\
 \frac{\partial\vect{X}}{\partial t}(\vect{s},t)
 &= \int_{\Omega_f}\vect{u}(\vect{x},t)
    \delta\!\left(\vect{x}-\vect{X}(\vect{s},t)\right)\,d\vect{x}.
 \label{eq:interpolate}
\end{align}
Here \(\rho\) and \(\mu\) are density and dynamic viscosity,
\(\vect{f}_{\mathrm{IB}}\) is the Eulerian force density, and \(\vect{F}\)
contains structural or constraint forces. The implementation also accounts for
prescribed deformation, rigid-body momentum, and hydrodynamic forces
~\cite{bhalla2013unified}. The Reynolds number
\begin{equation}
 \Rey=\frac{\rho U_{\mathrm{ref}}L_{\mathrm{ref}}}{\mu}
 \label{eq:reynolds}
\end{equation}
indexes the simulated operating conditions. In two dimensions, the scalar
vorticity is
\begin{equation}
 \Omega=\frac{\partial u_y}{\partial x}
       -\frac{\partial u_x}{\partial y}.
 \label{eq:vorticity}
\end{equation}

\subsection{Neural-Operator Surrogate Formulation}

Let \(t_{n+1}=t_n+\Delta t_n\). For the planar swimmer, the executed mapping is
\begin{equation}
 [\widehat u_x,\widehat u_y,\widehat\Omega,\widehat P]_{n+1}
 =
 \mathcal{G}^{2D}_{\theta}
 [u_x,u_y,\Omega,P,M_e,\Rey, \Delta t]_{n},
 \label{eq:planar-map}
\end{equation}
where \(M_e\) is a binary occupancy proxy and \(\Rey\) is broadcast over the
grid.

The volumetric models share the ordered input
\begin{align}
\vect{x}^{3D}_n=[
&u_x,u_y,u_z,\Omega_x,\Omega_y,\Omega_z,P,M_e, \Rey,\Delta t,\sin\phi,\cos\phi]_n ,
\label{eq:volumetric-input}
\end{align}
where \(\phi=2\pi t/T\) and the executed phase period is \(T=1\). Separate
parameter sets define
\begin{equation}
\begin{split}
 \mathcal{G}^{(u)}_{\theta}:\vect{x}^{3D}_n
 &\mapsto [u_x,u_y,u_z]_{n+1},\\
 \mathcal{G}^{(\omega)}_{\theta}:\vect{x}^{3D}_n
 &\mapsto [\Omega_x,\Omega_y,\Omega_z]_{n+1},\\
 \mathcal{G}^{(p)}_{\theta}:\vect{x}^{3D}_n
 &\mapsto P_{n+1}.
\end{split}
\label{eq:volumetric-maps}
\end{equation}
The three-dimensional formulation is therefore a suite of three
target-specific models, not one joint seven-output network. For the common
accuracy assessment, every prediction uses the reference field at \(t_n\);
this is teacher-forced evaluation in the standard temporal-modeling sense.

The stored fluid mask \(M_f\) is evaluated at the target time when defining
fluid-region errors because the swimmer moves. Both masks are grid-based
proxies derived from the immersed geometry; neither is an exact
sharp-interface solid volume.

\section{High-Fidelity Simulations and Data Preparation}
\label{sec:data}

IBAMR advances the fluid on an adaptive Cartesian hierarchy while storing the
swimmer with Lagrangian points. At a prescribed cadence, the workflow samples
Eulerian fields on a fixed Cartesian grid and writes one trajectory at each
Reynolds number. Cropping selects existing coordinate indices; it does not
interpolate or filter the retained fields. This regular representation supplies
the tensors required by the FNO while retaining physical coordinates,
trajectory-specific times, parameters, and geometry masks.

The planar database contains 50 trajectories with 201 frames and seven stored
fields on a \(256\times256\) crop. After skipping the first stored frame, the loader forms 200 temporal pairs per trajectory. Runs 0--44 provide 9,000
training pairs. Runs 45--49 provide 1000 held-out pairs at
\(\Rey=\{7300,7600,7900,8200,8500\}\). Because the training range ends at
\(\Rey=7000\), these are high-side extrapolation cases.

The volumetric database contains 50 trajectories, 101 retained frames, ten
stored fields, and a \(64^3\) crop. The physical time array has one row per
trajectory. Each trajectory supplies 100 temporal
pairs. Runs \(\{5,10,20,30,40\}\) are held out at
\(\Rey=\{750,1100,2200,4200,6200\}\); all other runs provide 4,500 training
pairs. All five held-out values lie within the training range
\(500\leq\Rey\leq8500\) and are classified as interpolation. Unequal source
frame counts were reconciled by retaining the largest common prefix by frame
count. This truncation leaves the trajectory-specific physical times intact
and is not temporal interpolation. The volumetric coordinates span
\(x\in[5.0,7.0]\),
\(y\in[-1.0,1.0]\), and
\(z\in[-0.5,0.5]\). Complete array layouts, dtypes, field order,
and validation checks are given in Appendix~\ref{app:data-schema}.

\begin{table*}[t]
\centering
\caption{High-fidelity simulation data used by the neural surrogates.}
\label{tab:datasets}
\small
\setlength{\tabcolsep}{3pt}
\begin{tabular}{p{0.11\textwidth}p{0.16\textwidth}p{0.10\textwidth}
                p{0.19\textwidth}p{0.24\textwidth}p{0.15\textwidth}}
\toprule
Case & Runs \(\times\) frames & Grid & Stored fields & Model inputs & Model targets \\
\midrule
Planar
& \(50\times201\)
& \(256^2\)
& \(\vect{u},\Omega,P,M_e,M_f\)
& \(\vect{u},\Omega,P,M_e,\Rey,\Delta t\)
& \(\vect{u},\Omega,P\) \\
Volumetric
& \(50\times104\)
& \(64^3\)
& \(\vect{u},\vect{\Omega},P,M_e,M_f\)
&  \(\vect{u},\vect{\Omega},P,M_e,\Rey,\Delta t, \phi\)
& \(\vect{u}\), or \(\vect{\Omega}\), or \(P\) \\
\bottomrule
\end{tabular}
\end{table*}

\section{Neural Operator and Experimental Protocol}
\label{sec:method}

\subsection{Fourier neural operator}

To create the surrogate model, the FNO has been used in the executed experiments.
A pointwise lifting layer maps the input to latent width \(d_v\). Each Fourier
block combines a local map with a truncated spectral convolution:
\begin{equation}
 \vect{v}_{\ell+1}
 =\sigma_\ell\!\left[
 W_\ell\vect{v}_\ell+
 \mathcal{M}_\ell\!\left\{
 \mathcal{F}^{-1}\!\left(
 R_\ell\odot\mathcal{F}(\vect{v}_\ell)\right)\right\}\right],
 \label{eq:fno-layer}
\end{equation}
where \(R_\ell\) contains learned complex multipliers on retained modes,
\(W_\ell\) is a pointwise linear map, and \(\mathcal{M}_\ell\) is a pointwise
MLP. In the executed four-block networks, \(\sigma_\ell\) is GELU after the
first three blocks and the identity after the fourth. The blocks are followed
by a pointwise projection. 

The planar model appends two internally constructed grid coordinates to its six
data channels. Each volumetric model receives three coordinates explicitly
among its 15 channels. Table~\ref{tab:architectures} reports the executed
architectures. The parameter count treats each complex coefficient as two real
scalars. The planar checkpoint contains 2,111,300 PyTorch parameter elements,
of which 2,097,152 are complex; it therefore represents 4,208,452 real-scalar
degrees of freedom.

\begin{table*}[t]
\centering
\caption{Executed neural-operator architectures.}
\label{tab:architectures}
\small
\resizebox{\textwidth}{!}{%
\begin{tabular}{llllllllr}
\toprule
Case and target & Model & Modes & Width & Layers & Padding & Input & Output
& Parameters \\
\midrule
Planar, joint fields & FNO2d & \(16^2\) & 32 & 4 & 8 & 6 & 4 & 4,208,452 \\
Volumetric, velocity & FNO3d & \(24^3\) & 24 & 4 & 8 & 15 & 3 & 254,814,243 \\
Volumetric, vorticity & FNO3d & \(24^3\) & 24 & 4 & 8 & 15 & 3 & 254,814,243 \\
Volumetric, pressure & FNO3d & \(24^3\) & 24 & 4 & 8 & 15 & 1 & 254,814,049 \\
\bottomrule
\end{tabular}}
\end{table*}

\subsection{Optimization, normalization, and model selection}

The planar model was trained for 100 epochs with Adam, initial learning rate
\(10^{-3}\), weight decay \(10^{-4}\), batch size 8, and per-batch cosine
annealing. Its training loss is the mean per-sample joint relative \(L^2\) norm
of the four decoded physical outputs. Channelwise Gaussian normalizers were
fit on the 8,955 training pairs according to the log and saved normalizer
statistics. 
Selection minimized the held-out mean
per-sample joint full-domain relative \(L^2\) loss, whose value is 0.0351195.

The 3D example uses Adam with the same initial learning
rate and weight decay, per-step cosine annealing, batch size 25, and seed 0.
Their loss is the mean per-sample joint relative \(L^2\) norm in the normalized
target space for the corresponding target group. Inputs and outputs use fixed
channelwise affine scaling,
\begin{equation}
 q_{\mathrm{scaled}}=\frac{q-c_q}{s_q},
 \label{eq:affine-scaling}
\end{equation}
with constants embedded in the configuration and checkpoint. All decoded-field accuracy metrics are computed after decoding to stored
physical numerical units. The volumetric runs were configured for up to 1,000 epochs.  

\subsection{Error definitions and physical diagnostics}

For channels \(\mathcal{C}\) and spatial region \(\mathcal{R}\), the global
energy-weighted relative error is
\begin{equation}
 E_{\mathrm{global}}^{(\mathcal{R},\mathcal{C})}
 =
 \left[
 \frac{\sum_{b,\vect{x}\in\mathcal{R},c\in\mathcal{C}}
 (\widehat Y_{b,c}(\vect{x})-Y_{b,c}(\vect{x}))^2}
 {\sum_{b,\vect{x}\in\mathcal{R},c\in\mathcal{C}}
 Y_{b,c}(\vect{x})^2}
 \right]^{1/2}.
 \label{eq:global-relative}
\end{equation}

The per-sample relative error applies the same ratio to one sample; reported means,
standard deviations, medians, and quartiles summarize samples rather than
random seeds. MSE, RMSE, and MAE are accumulated pointwise after physical
decoding. Vector-group errors combine their three components, whereas pressure
is always reported separately. Fluid-region metrics use \(M_f\) at the target
time. Pressure is primarily evaluated in the solver-exported gauge. A secondary
gauge-insensitive diagnostic subtracts the spatial mean from prediction and
target separately for each sample and region. For the planar physical
diagnostics, velocity derivatives use actual coordinate spacing, central
differences in the interior, and one-sided differences at nonperiodic
boundaries. Divergence and the consistency between predicted vorticity and the
curl of predicted velocity are evaluated over the target-time fluid mask.

\section{Results}
\label{sec:results}

\subsection{Two-Dimensional Swimmer}
\label{sec:results-planar}

\begin{figure}[!htbp]
\centering
\includegraphics[width=0.96\textwidth]{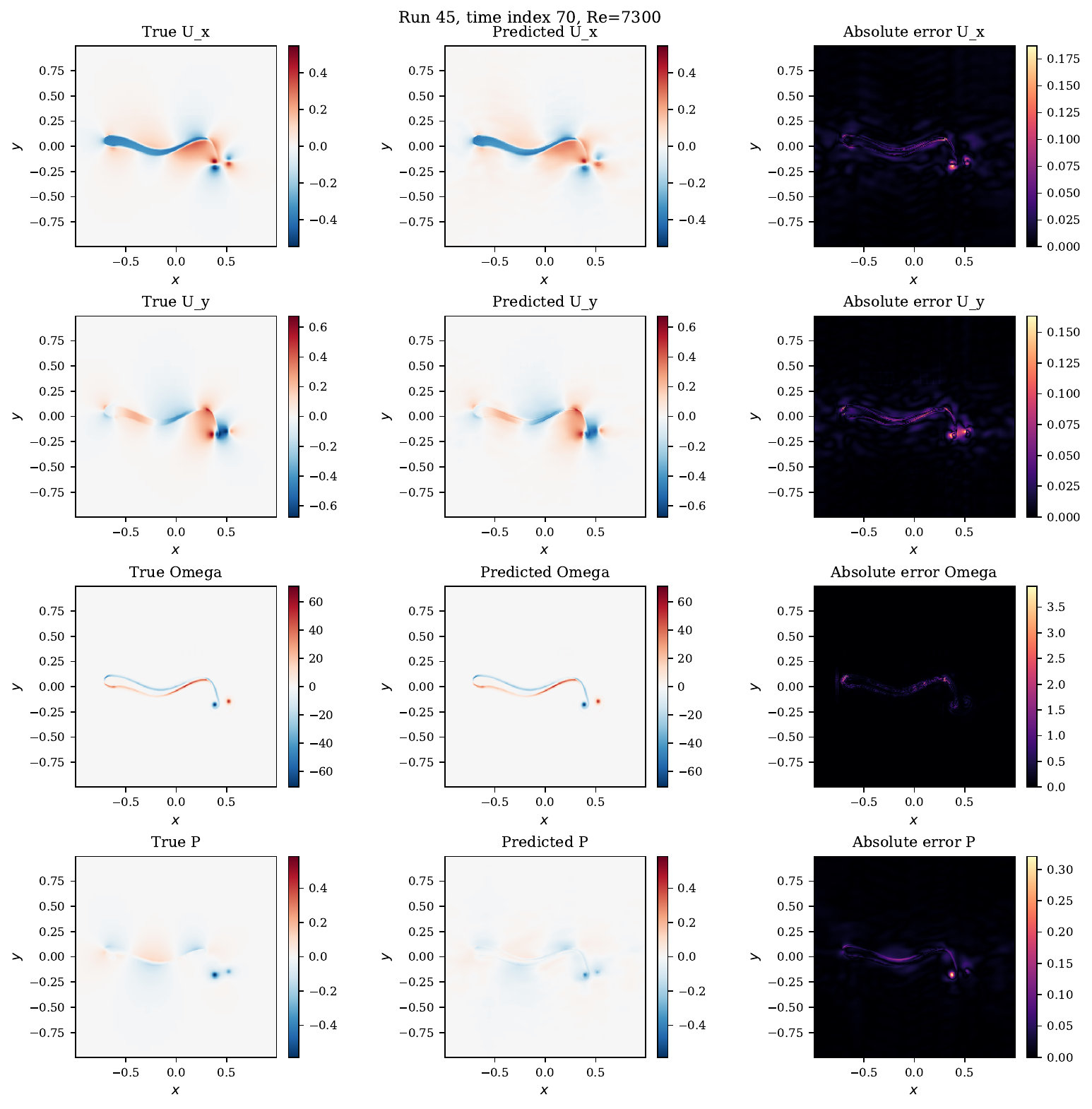}
\caption{Planar swimmer field prediction for held-out run 45
(\(\Rey=7300\)), from retained input time index 70 (\(t=0.71\)) to target
time \(t=0.72\).
Rows show \(u_x\), \(u_y\), \(\Omega\), and \(P\); columns show the
immersed-boundary target, FNO prediction, and absolute error. Target and
prediction use the same color limits within each row.}
\label{fig:planar-prediction}
\end{figure}

Figure~\ref{fig:planar-prediction} compares the reference and prediction for a
held-out planar sample. Foo all 1000
held-out temporal pairs. The full-domain global relative error is \(3.57\%\). 

The joint error increases monotonically across the five held-out Reynolds
numbers (Table~\ref{tab:planar-re}). Every row represents 200 temporal pairs
from one trajectory. These cases test high-side extrapolation relative to the
training range, not interpolation within the complete generated sweep.

\begin{table}[t]
\centering
\caption{Planar swimmer error by held-out Reynolds number. Values are global
joint relative \(L^2\) percentages.}
\label{tab:planar-re}
\begin{tabular}{rrrrl}
\toprule
Run & \(\Rey\) & Full [\%] & Fluid [\%] & Regime \\
\midrule
45 & 7300 & 3.371 & 3.033 & extrapolation \\
46 & 7600 & 3.460 & 3.121 & extrapolation \\
47 & 7900 & 3.568 & 3.235 & extrapolation \\
48 & 8200 & 3.654 & 3.311 & extrapolation \\
49 & 8500 & 3.784 & 3.438 & extrapolation \\
\bottomrule
\end{tabular}
\end{table}

The data-only training objective does not enforce incompressibility or couple
the independently predicted vorticity to the velocity curl. Table
~\ref{tab:planar-physics} shows that the prediction has larger divergence and
vorticity inconsistency than the target fields. These discrepancies motivate
physics-aware objectives even though the field-wise error is small.

\begin{table}[t]
\centering
\caption{Planar swimmer physical diagnostics over the target-time fluid mask.
Derivatives use nonperiodic central interior and one-sided boundary stencils.}
\label{tab:planar-physics}
\small
\begin{tabular}{lrrr}
\toprule
Diagnostic & Target & Prediction & Discrepancy \\
\midrule
Mean absolute divergence & 0.04059 & 0.17703 & 0.17047 \\
RMS divergence & 0.21803 & 0.54747 & 0.56037 \\
Mean absolute vorticity inconsistency & 0.00055 & 0.19069 & 0.19060 \\
RMS vorticity inconsistency & 0.02607 & 0.70147 & 0.70184 \\
\bottomrule
\end{tabular}
\end{table}

\subsection{Three-Dimensional Swimmer}
\label{sec:results-volumetric}

\begin{figure}[!htbp]
\centering
\includegraphics[width=0.98\textwidth]{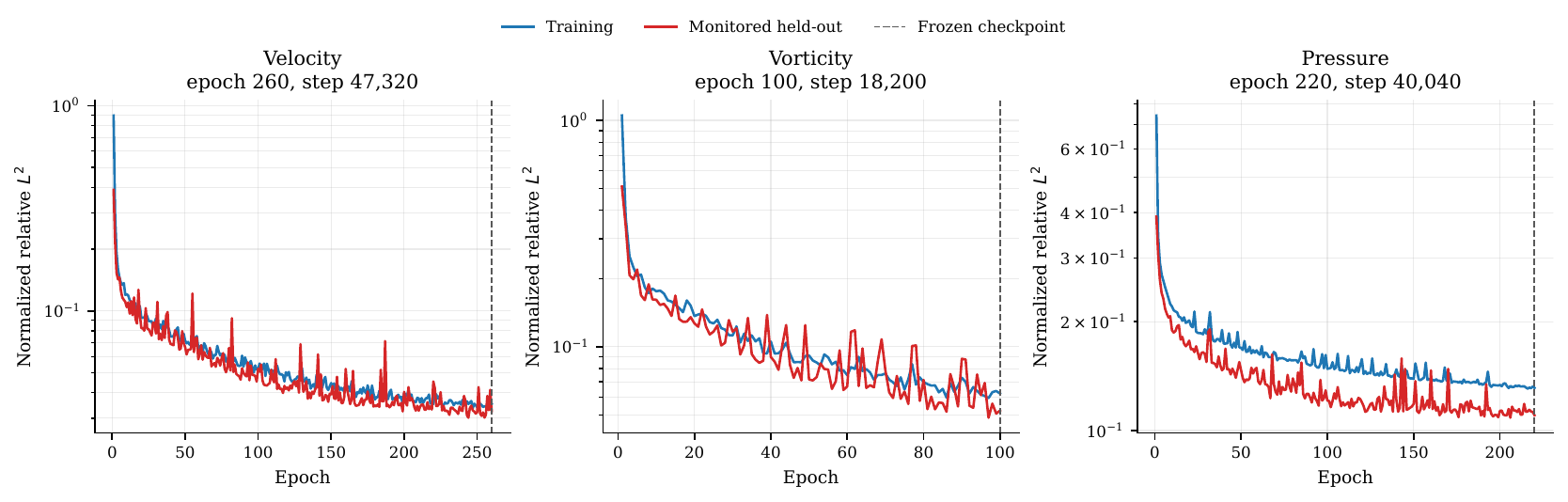}
\caption{Training and monitored held-out relative \(L^2\) histories in
normalized target space for the three frozen volumetric models. Dashed lines
mark the actual saved checkpoints: velocity epoch 260, vorticity epoch 100, and
pressure epoch 220.}
\label{fig:volumetric-history}
\end{figure}

Figure~\ref{fig:volumetric-history} shows the training and monitored
held-out losses through the epochs. Table~\ref{tab:volumetric-groups} summarizes the three target groups over all
500 held-out temporal pairs. Velocity achieves a \(3.44\%\) full-domain global
relative error, vorticity \(5.59\%\), and pressure \(19.20\%\). The corresponding
mean sample errors are \(3.08\%\), \(5.18\%\), and \(11.02\%\). Pressure has a
strongly skewed sample distribution: its median is \(3.88\%\), whereas a small
number of high-relative-error samples increase the mean and standard deviation.
Spatial-mean removal changes the full-domain pressure error only from
\(19.203\%\) to \(19.213\%\), indicating that a uniform gauge offset is not the
dominant source of error.

\begin{table*}[t]
\centering
\caption{Target-group errors for the 3D swimmer. Global errors are
energy-weighted ratios; mean, standard deviation, and median summarize 500
per-sample joint ratios. RMSE and MAE
are joint pointwise errors in decoded stored simulation units.}
\label{tab:volumetric-groups}
\small
\resizebox{\textwidth}{!}{%
\begin{tabular}{lrrrrrr}
\toprule
Target group & Full global [\%] & Fluid global [\%]
& Sample mean \(\pm\) std. [\%] & Sample median [\%] & RMSE & MAE \\
\midrule
Velocity & 3.442 & 3.920 & \(3.077\pm1.081\) & 2.824
& \(3.8070{\times}10^{-4}\) & \(9.8721{\times}10^{-5}\) \\
Vorticity & 5.586 & 6.354 & \(5.184\pm1.405\) & 4.905
& \(1.9834{\times}10^{-2}\) & \(3.3474{\times}10^{-3}\) \\
Pressure & 19.203 & 18.352 & \(11.022\pm33.780\) & 3.881
& \(6.8940{\times}10^{-4}\) & \(8.2131{\times}10^{-5}\) \\
\bottomrule
\end{tabular}}
\end{table*}

The component-wise values in Table~\ref{tab:volumetric-components} show a
narrow velocity range of \(3.24\%\)--\(4.01\%\). The transverse vorticity
component \(\Omega_y\) is the most difficult vortical output at \(7.94\%\),
compared with \(5.37\%\) for \(\Omega_x\) and \(4.66\%\) for \(\Omega_z\).

\begin{table*}[t]
\centering
\caption{Component-wise volumetric errors in decoded stored simulation units.
Relative \(L^2\) values are global percentages.}
\label{tab:volumetric-components}
\small
\begin{tabular}{lrrrr}
\toprule
Field & Full [\%] & Fluid [\%] & RMSE & MAE \\
\midrule
\(u_x\) & 3.542 & 4.339 & \(3.7248{\times}10^{-4}\) & \(1.1001{\times}10^{-4}\) \\
\(u_y\) & 3.240 & 3.696 & \(4.6752{\times}10^{-4}\) & \(1.0408{\times}10^{-4}\) \\
\(u_z\) & 4.011 & 3.992 & \(2.7836{\times}10^{-4}\) & \(8.2081{\times}10^{-5}\) \\
\(\Omega_x\) & 5.371 & 6.021 & \(2.3804{\times}10^{-2}\) & \(3.6514{\times}10^{-3}\) \\
\(\Omega_y\) & 7.943 & 7.913 & \(1.8258{\times}10^{-2}\) & \(3.9127{\times}10^{-3}\) \\
\(\Omega_z\) & 4.660 & 5.856 & \(1.6739{\times}10^{-2}\) & \(2.4780{\times}10^{-3}\) \\
\(P\) & 19.203 & 18.352 & \(6.8940{\times}10^{-4}\) & \(8.2131{\times}10^{-5}\) \\
\bottomrule
\end{tabular}
\end{table*}

\begin{figure}[!htbp]
\centering
\includegraphics[width=0.80\textwidth]{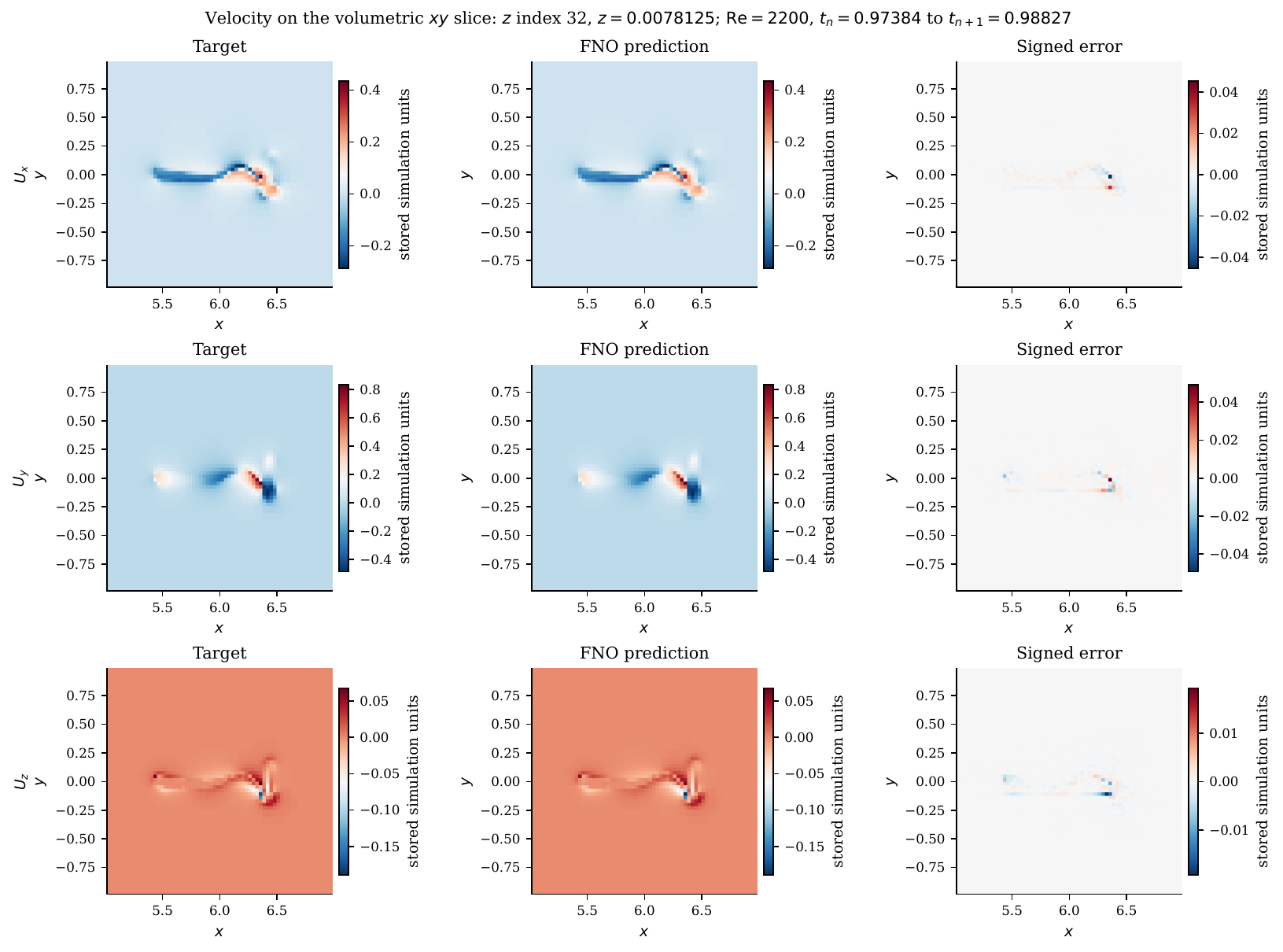}
\caption{Velocity prediction on an \(xy\) slice through the volumetric
swimmer field. Rows show \(u_x,u_y,u_z\); columns show target, FNO prediction,
and signed error.}
\label{fig:volumetric-velocity}
\end{figure}

\begin{figure}[!htbp]
\centering
\includegraphics[width=0.80\textwidth]{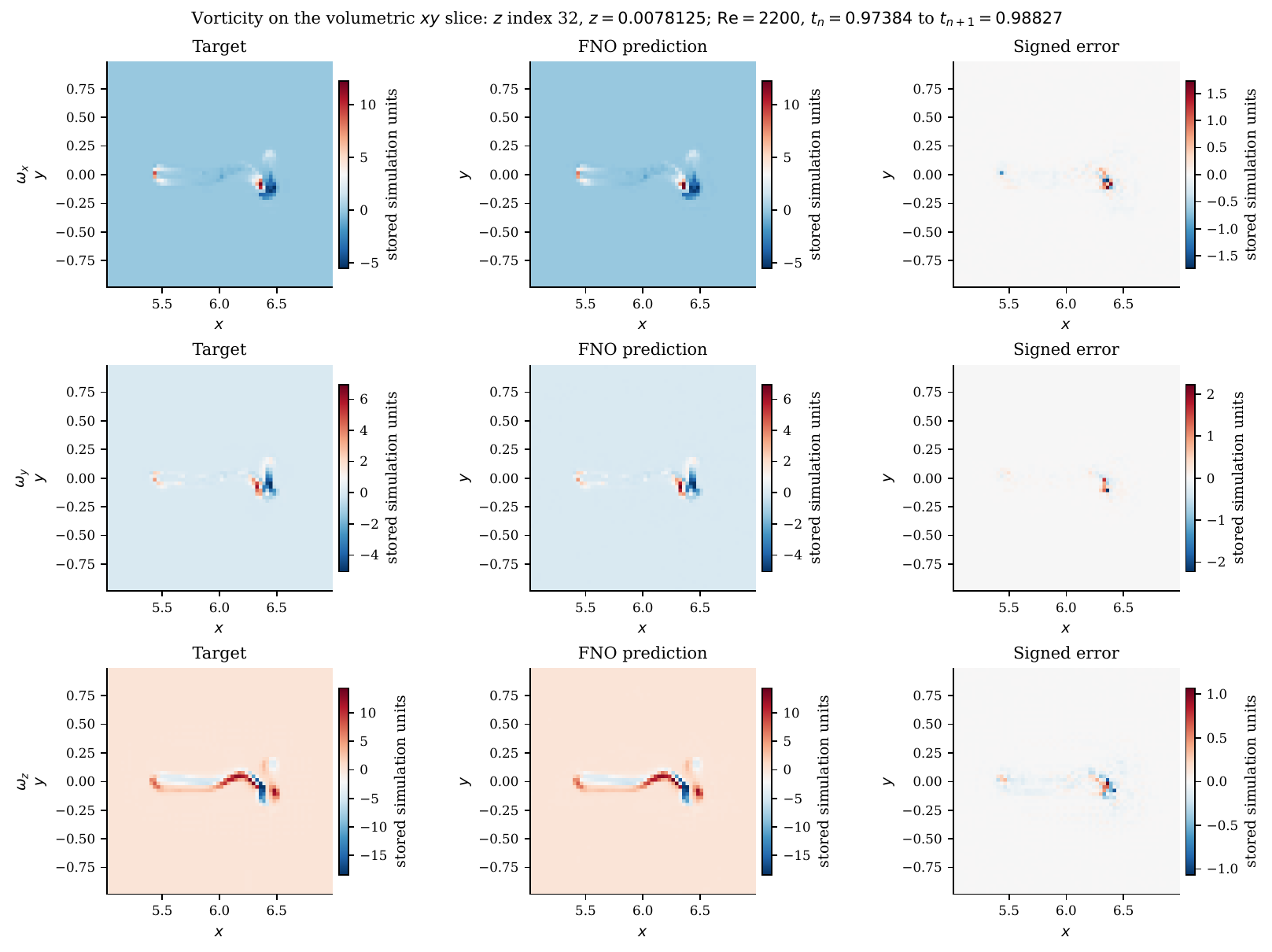}
\caption{Vorticity prediction on the same \(xy\) slice through the volumetric
field. Rows show \(\Omega_x,\Omega_y,\Omega_z\); columns show target, FNO
prediction, and signed error.}
\label{fig:volumetric-vorticity}
\end{figure}

\begin{figure}[!htbp]
\centering
\includegraphics[width=0.80\textwidth]{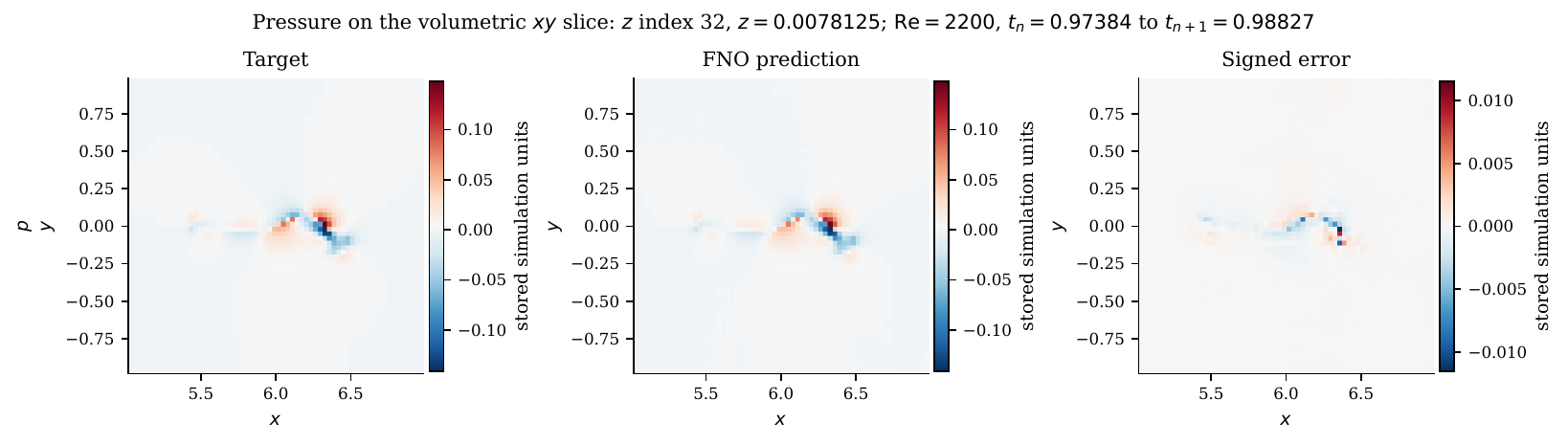}
\caption{Pressure target, FNO prediction, and signed error on the same
\(xy\) slice through the volumetric field. Pressure is shown in the stored
solver gauge.}
\label{fig:volumetric-pressure}
\end{figure}

Figures~\ref{fig:volumetric-velocity}--\ref{fig:volumetric-pressure} show a
common \(xy\) slice through genuine volumetric predictions. The displayed
sample is held-out run 20 at \(\Rey=2200\), from input frame 52
(\(t=0.97384\)) to target frame 53 (\(t=0.98827\)). The slice is stored
\(z\)-index 32, corresponding to \(z=0.0078125\). Target and prediction share
the same limits for each component; signed error uses a separate symmetric
scale.

\begin{figure}[!htbp]
\centering
\includegraphics[width=0.98\textwidth]{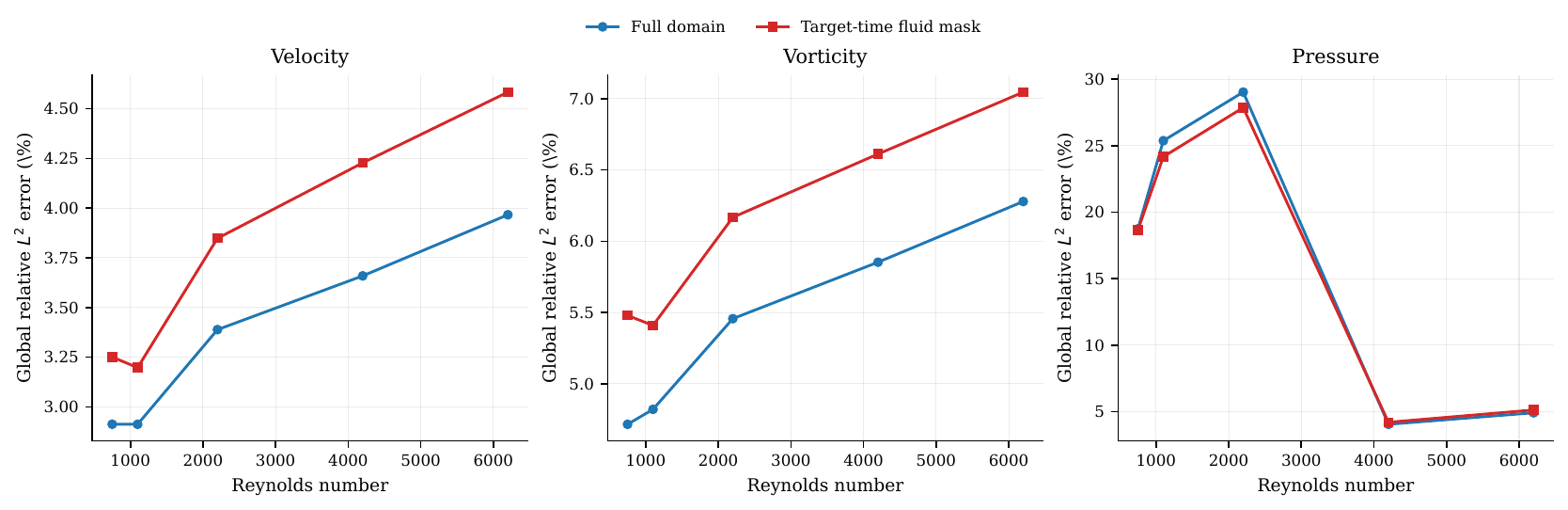}
\caption{Global relative \(L^2\) error versus Reynolds number for the
target-specific volumetric models. Full-domain and target-time fluid-mask
ratios are reported separately. Each point represents one held-out trajectory,
not an uncertainty estimate across seeds.}
\label{fig:volumetric-re}
\end{figure}

Velocity and vorticity errors generally increase across the held-out
within-range Reynolds numbers (Table~\ref{tab:volumetric-re} and
Figure~\ref{fig:volumetric-re}). Pressure behaves nonmonotonically: its
relative error is highest at \(\Rey=2200\) and much lower at 4200 and 6200.

\begin{table}[t]
\centering
\caption{Full-domain global relative \(L^2\) error by held-out volumetric
trajectory. Each row contains 101 temporal pairs and is an interpolation case.}
\label{tab:volumetric-re}
\small
\begin{tabular}{rrrrrr}
\toprule
Run & \(\Rey\) & Samples & Velocity [\%] & Vorticity [\%] & Pressure [\%] \\
\midrule
5  & 750  & 100 & 2.913 & 4.716 & 18.751 \\
10 & 1100 & 100 & 2.913 & 4.821 & 25.377 \\
20 & 2200 & 100 & 3.389 & 5.457 & 29.030 \\
30 & 4200 & 100 & 3.659 & 5.853 & 4.057 \\
40 & 6200 & 100 & 3.966 & 6.279 & 4.921 \\
\bottomrule
\end{tabular}
\end{table}

\subsection{Computational record}

Table~\ref{tab:computational} reports synchronized pure-forward timing. The
volumetric measurements use two warm-up calls followed by 500 batch-one calls
per model; data loading, host-to-device transfer, decoding, and metric
accumulation are excluded. The planar timing was recorded in its original
experiment at batch size 8. 

\begin{table*}[t]
\centering
\caption{Neural-model computational record. Peak allocation is unavailable in
the planar experiment record. Timing is not compared with the immersed-boundary solver
and therefore does not establish solver speedup.}
\label{tab:computational}
\small
\resizebox{\textwidth}{!}{%
\begin{tabular}{lrrrrl}
\toprule
Model & Parameters & Batch & Forward [ms/sample] & Peak allocated [GB] & GPU \\
\midrule
2D joint FNO & 4,208,452 & 8 & 1.756 & -- & NVIDIA A100 40 GB \\
3D velocity FNO & 254,814,243 & 1 & 26.678 & 1.284 & RTX 4000 Ada 20 GB \\
3D vorticity FNO & 254,814,243 & 1 & 26.670 & 1.284 & RTX 4000 Ada 20 GB \\
3D pressure FNO & 254,814,049 & 1 & 26.756 & 1.282 & RTX 4000 Ada 20 GB \\
\bottomrule
\end{tabular}}
\end{table*}

\subsection{Planar and volumetric comparison}

Both dimensions support field-resolved temporal prediction, but they are not a
controlled architecture ablation. The planar held-out set extrapolates beyond
the training Reynolds-number range, whereas the volumetric set samples
within-range conditions. The resolutions, channel maps, normalization, model
sizes, and joint-versus-separate target strategies also differ. Within these
constraints, the shared qualitative result is clear: velocity and vorticity
are represented more accurately than pressure, and global field agreement
does not by itself guarantee differential consistency or recursive stability.

\section{Discussion}
\label{sec:discussion}

\subsection{Accuracy across hydrodynamic fields}

The neural operators reproduce the dominant swimmer-generated flow and wake
structures in both dimensions. In the planar case, scalar vorticity has the
smallest relative error despite its larger numerical amplitude, while pressure
has the largest. The volumetric models show a similar ordering at the
target-group level. Separating the three volumetric target groups avoids direct
competition between variables of different scales, but it does not enforce
cross-model identities such as
\(\vect{\Omega}=\nabla\times\vect{u}\). Physics-based coupling remains an
important opportunity.

The pressure results deserve particular care. A global pressure error of
\(19.2\%\) in the volumetric case is substantially larger than the velocity and
vorticity group errors. Spatial-mean removal does not improve it, so the error
cannot be attributed primarily to a uniform gauge offset. The pronounced
Reynolds-number and sample dependence suggests that future pressure models
should examine near-body gradients, force-relevant integrals, and
regime-balanced objectives in addition to global norms.

\subsection{Generalization and dimensionality}

The planar joint error increases gradually over high-side Reynolds-number
extrapolation from 7300 to 8500. Volumetric velocity and vorticity errors also
increase over most of the within-range held-out values, while pressure is
nonmonotonic. These trends support parameter-conditioned field modeling but do
not establish generalization to a new gait, swimmer geometry, or material.

The volumetric models contain roughly 255 million real-scalar parameters each,
compared with 4.2 million for the planar network. This increase reflects
three-dimensional spectral weights and the executed mode/width choices, not a
claim that such a parameter ratio is intrinsically necessary. More economical
factorized, geometry-aware, or multiresolution operators should be compared
under the same splits and metrics.

\subsection{Physical and engineering implications}

The planar divergence and vorticity-consistency diagnostics show that visually
and globally accurate fields may still violate differential relationships.
For engineering analysis, future evaluations should include surface- or
interface-sensitive errors and derived loads where the immersed-boundary data
support them. The present masks provide useful geometry conditioning and
fluid-region reporting, but a signed-distance field or Lagrangian geometry
encoder could better resolve subcell interface location.

\subsection{Limitations}

All results are generated from simulation with prescribed swimmer kinematics,
and the surrogate has not yet been validated against experimental measurements.
The reported accuracy primarily concerns one-step prediction using flow fields at the previous time step as inputs.
The present parameter study primarily assesses generalization with respect to
Reynolds number and does not establish transfer to unseen swimmer geometries,
gaits, or actuation patterns. Finally, the experiments use a single training
seed, and pressure is evaluated in the exported solver gauge, with a
mean-removed secondary diagnostic for the volumetric case.

\subsection{Future work}

A stronger protocol should reserve trajectory-level validation and final test
sets, fit all data-dependent transformations on training trajectories, and
report multiple independent seeds. Methodologically, physics-aware losses,
cross-field consistency, signed-distance geometry channels, factorized
volumetric operators, and direct temporal models are promising. Matched
solver-surrogate timing and derived-load comparisons are needed for engineering
cost claims. New gaits, geometries, and experimental measurements are required
before extending the conclusions beyond the present simulation family.

\section{Conclusion}
\label{sec:conclusion}

We developed Fourier neural-operator surrogates for hydrodynamic field
prediction around immersed-boundary eel swimmers in two and three dimensions.
The planar model jointly predicts velocity, scalar vorticity, and pressure,
with full-domain global relative error of \(3.51\%\) on held-out high-side Reynolds-number trajectories. Three
target-specific volumetric models predict velocity, vorticity, and pressure
with corresponding group errors of \(3.44\%\), \(5.59\%\), and \(19.2\%\) on
held-out within-range trajectories.

These results position immersed-boundary swimmer locomotion as a substantive
engineering application for neural operators. They also identify clear
research needs: more accurate pressure, stronger differential consistency,
validation-based model selection, stable temporal prediction, and matched
computational-cost studies. Within those limitations, the frozen models and
reproducible evaluation artifacts provide a concrete basis for improving
neural surrogates for moving-boundary fluid-structure interaction.

\section*{Data and Code Availability}

The code used for high-fidelity simulation, dataset generation, neural-operator
training, evaluation, and post-processing will be made publicly available soon at

\begin{center}
\url{https://github.com/eshaghi-ms/NOforSwimmers}.
\end{center}

The repository will contain the implementations and supporting files required to
reproduce the planar and volumetric swimmer studies presented in this work,
including training configurations, model definitions, evaluation scripts, and
post-processing routines. Links to the corresponding datasets will also be
provided through the repository.

\section*{Acknowledgments}

The authors would like to acknowledge the support provided by the German Academic Exchange Service (DAAD) through a scholarship awarded to Mohammad Sadegh Eshaghi during this research, as well as the Compute Servers of TU Ilmenau for providing computational resources.

\appendix

\section{Dataset Schema and Channel Order}
\label{app:data-schema}

The planar numerical array has layout
\([r,t,c,y,x]=[50,201,7,256,256]\) and dtype \texttt{float32}. Its ordered
fields are
\[
[u_x,u_y,\Omega,P,\nabla\!\cdot\vect{u},M_e,M_f].
\]
The regular-grid files separate
\path{data.npy}, \path{time.npy}, \path{x.npy}, \path{y.npy},
\path{field_names.npy}, \path{params.npy}, and run identifiers so that the
field data can be memory mapped.

The volumetric array
\path{examples/ConstraintIB/ibamr_eel3d_runs/dataset/data.npy} has layout
\[
[r,t,c,z,y,x]=[50,104,10,64,64,64]
\]
and dtype \texttt{float32}. Its ordered fields are
\[
[u_x,u_y,u_z,\Omega_x,\Omega_y,\Omega_z,P,
 \nabla\!\cdot\vect{u},M_e,M_f].
\]
The time array has shape \([50,104]\), dtype \texttt{float64}, and range
\(0\)--\(2.0\); each trajectory is strictly increasing. Parameter
metadata uses a structured \texttt{float64} dtype. The common-prefix rule operates on
frame count and does not modify the retained time values.

All three volumetric models use the input order
\[
[u_x,u_y,u_z,\Omega_x,\Omega_y,\Omega_z,P,M_e,\Rey,\Delta t,
\sin\phi,\cos\phi].
\]

Each output uses the entries corresponding to its ordered target fields.
The planar split is runs 0--44 for training and 45--49 for held-out
selection/evaluation. The volumetric held-out runs are
\([5,10,20,30,40]\); the training set is the complement in runs 0--49.

\bibliographystyle{unsrtnat}
\bibliography{references}

\end{document}